\documentclass[runningheads]{llncs}
\usepackage{graphicx}

\usepackage{tikz}
\usepackage{comment}
\usepackage{amsmath,amssymb} 
\usepackage{color}
\usepackage{booktabs}
\usepackage{multirow}
\usepackage[pagebackref,breaklinks,colorlinks]{hyperref}
\usepackage[accsupp]{axessibility}  

\newcommand{\gray}[1]{\textcolor{lightgray}{#1}}

\usepackage[width=122mm,left=12mm,paperwidth=146mm,height=193mm,top=12mm,paperheight=217mm]{geometry}

\begin{document}
\pagestyle{headings}
\mainmatter
\def\ECCVSubNumber{168}  

\title{In Defense of Online Models for Video Instance Segmentation} 

\titlerunning{In Defense of Online Models for VIS}
%
\author{Junfeng Wu\inst{1}* ,
Qihao Liu\inst{2}*,
Yi Jiang\inst{3},
Song Bai\inst{3\dag}, 
Alan Yuille\inst{2},
Xiang Bai\inst{1}
}
\authorrunning{J. Wu, Q. Liu et al.}
%
\institute{
$^{1}$ Huazhong University of Science and Technology \\
$^{2}$ Johns Hopkins University\qquad  $^{3}$ ByteDance 
}
\maketitle
\def\thefootnote{*}\footnotetext{First two authors contributed equally. Work done during an internship at ByteDance.}
\def\thefootnote{\dag}\footnotetext{Corresponding author:
\href{mailto:songbai.site@gmail.com}{\color{black}{songbai.site@gmail.com}}
}\def\thefootnote{\arabic{footnote}}
\begin{sloppypar} 
\begin{abstract}

In recent years, video instance segmentation (VIS) has been largely advanced by offline models, while online models gradually attracted less attention possibly due to their inferior performance. However, online methods have their inherent advantage in handling long video sequences and ongoing videos while offline models fail due to the limit of computational resources. Therefore, it would be highly desirable if online models can achieve comparable or even better performance than offline models. By dissecting current online models and offline models, we demonstrate that the main cause of the performance gap is the error-prone association between frames caused by the similar appearance among different instances in the feature space. Observing this, we propose an online framework based on contrastive learning that is able to learn more discriminative instance embeddings for association and fully exploit history information for stability. Despite its simplicity, our method outperforms all online and offline methods on three benchmarks. Specifically, we achieve 49.5 AP on YouTube-VIS 2019, a significant improvement of 13.2 AP and 2.1 AP over the prior online and offline art, respectively. Moreover, we achieve 30.2 AP on OVIS, a more challenging dataset with significant crowding and occlusions, surpassing the prior art by 14.8 AP. 
The proposed method won \textbf{first place} in the video instance segmentation track of the 4th Large-scale Video Object Segmentation Challenge (CVPR2022).
We hope the simplicity and effectiveness of our method, as well as our insight on current methods, could shed light on the exploration of VIS models. The code is available at \href{https://github.com/wjf5203/VNext}{https://github.com/wjf5203/VNext}.

\keywords{Video instance segmentation, Online model, Contrastive learning}
\end{abstract}
 \end{sloppypar}
\section{Introduction}

Video instance segmentation aims at detecting, segmenting, and tracking object instances simultaneously in a given video. It attracted considerable attention after first defined~\cite{MaskTrackRCNN} in 2019 due to the huge challenge and the wide applications in video understanding, video editing, autonomous driving, augmented reality, etc. Current VIS methods can be categorized as online or offline methods. Online methods~\cite{MaskTrackRCNN,sipmask,compfeat,STMask,CrossVIS,PCAN} take as input a video frame by frame, detecting and segmenting objects per frame while tracking instances and optimizing results across frames. Offline methods~\cite{MaskProp,ProposeReduce,STEmSEG,VisTR,IFC,seqformer}, in contrast, take the whole video as input and generate the instance sequence of the entire video with a single step.

\begin{figure}[t]
	\centering
	\includegraphics[width=0.95\textwidth]{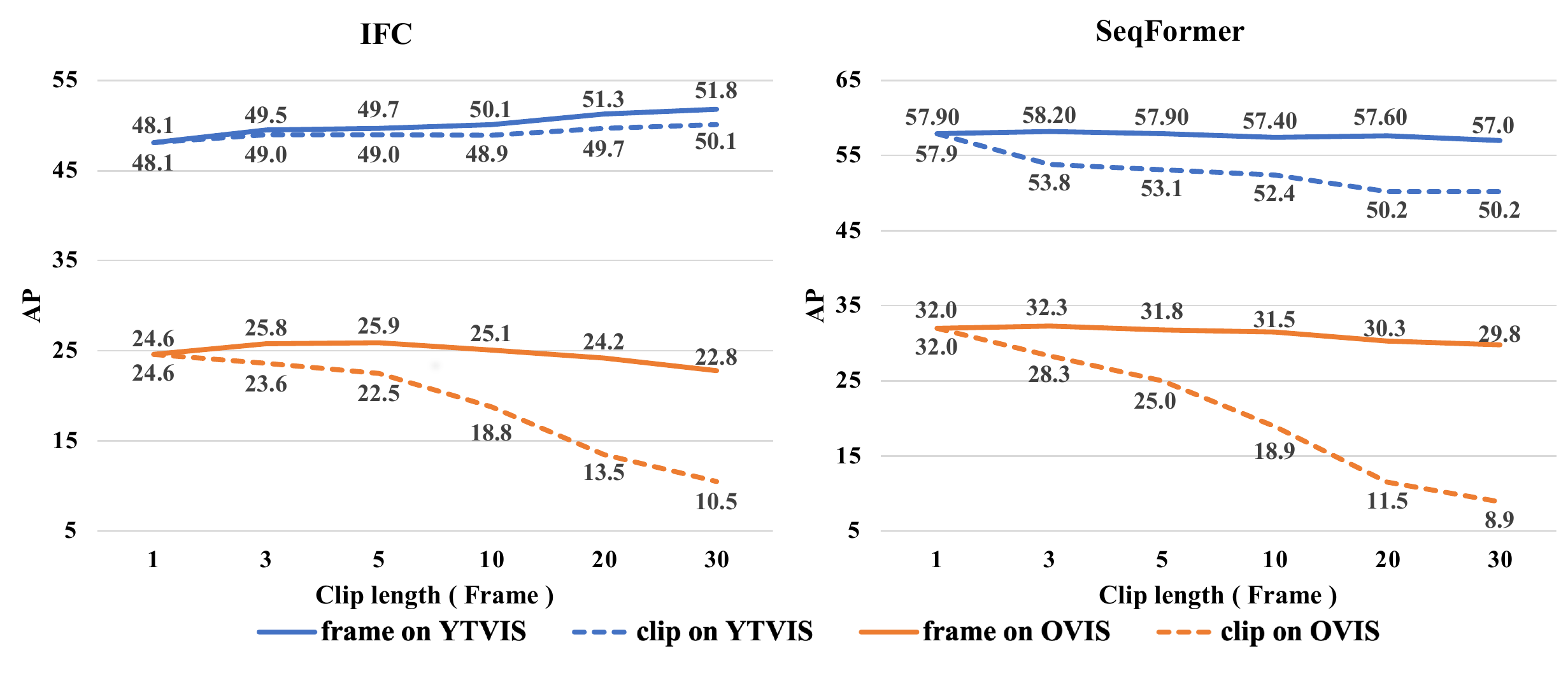}
	\caption{\textbf{Oracle experiments on SOTA offline methods: } We analyze current VIS methods and visualize some results here. More results can be found in Sec.~\ref{sec:VIS-Analysis}. `YTVIS' is short for YouTube-VIS 2019. `Frame' and `clip' stand for frame oracle and clip oracle experiments, respectively. For frame oracles, we provide the ground-truth instance ID both within each clip and between adjacent clips. Thus the performance only depends on the quality of the estimated segmentation masks. For clip oracles, we only provide the ground-truth instance ID between adjacent clips, and the method is required to do association within the clips by itself. Therefore the gaps between frame oracles and clip oracles show the effect of the black-box association done within current offline models. When clip length is 1, the method is doing per-frame segmentation.}
	\label{fig:motivation}
\end{figure}

Despite high performance, the requirement of full videos limits the application scenarios of offline models, especially for scenarios involving long video sequences (videos exceed 50 frames on GPU of 32G RAM~\cite{seqformer}) and ongoing videos (\eg, videos in autonomous driving and augmented reality). However, online models are usually inferior to the contemporaneous offline models by over 10 AP, which is a huge drawback. Previously, little work tries to explain the performance gap between these two paradigms, or gives insight into the high performance of the offline paradigm. A common attempt of the latter is made from the inherent advantages of offline models being able to skip the error-accumulating tracking steps~\cite{ProposeReduce} and utilize the richer information provided by multiple frames to improve segmentation~\cite{IFC,STEmSEG,seqformer,efficient}. However, does that really explain the high performance of current offline methods? What's the main problem causing the poor performance of online models? Can online models achieve performance comparable to, or even better than, SOTA offline ones?

To deeply understand the performance of both online and offline models, we analyze in detail three SOTA methods (offline: IFC~\cite{IFC} and SeqFormer~\cite{seqformer}, online: CrossVIS~\cite{CrossVIS}) on two datasets (YouTube-VIS~\cite{MaskTrackRCNN} and OVIS~\cite{ovis}) that have different difficulty levels. `Simple video' refers to the video in YouTube-VIS. These videos are much shorter and only contain very slight occlusions, simple motions, and smooth changes in illumination and object shapes. `Complex video' refers to the video in OVIS. Please see Sec.~\ref{sec:dataset} for details. The results (Fig.~\ref{fig:motivation}) of our oracle experiments give us a deep understanding of current SOTA methods:

From the perspective of instance segmentation, per-clip segmentation doesn't outperform per-frame segmentation a lot in mask quality, and mask quality is also not the reason for the poor performance of online methods: CrossVIS even outperforms its contemporaneous work (\ie IFC) in frame oracle experiments on both datasets (results in Table~\ref{table:crossvis_IFC}). What's more, per-clip segmentation of current SOTA methods is not effective and robust. Multiple frames do provide more information and improve the mask quality by 3.7 AP for IFC on YouTube-VIS (Fig.~\ref{fig:motivation}). But it only works for some cases: per-clip segmentation doesn't improve the performance of SeqFormer a lot. In addition, when testing on more challenging datasets like OVIS, segmentation on multiple frames even degrades the performance by 1.8 and 2.2 AP on IFC and SeqFormer respectively when clip size becomes longer.
Although in theory, per-clip segmentation has its inherent advantage to using multiple frames, it still requires further exploration, especially in how to utilize information in multiple frames and how to handle complex motion patterns, occlusion, and object deformation. Currently, we don't see an obvious gap between the mask qualities of per-clip and per-frame segmentation. 

From the perspective of association, a huge advantage of the offline methods is their ability to avoid the use of hand-designed association modules. It works well on simple cases of the YouTube-VIS dataset. We demonstrate that it is the main reason causing the performance gap between the current online and offline paradigms. However, this black-box association process done within offline models also gets worse rapidly when the video becomes complex (degrades the performance of IFC by 12.3 AP and SeqFormer by 20.9 AP on OVIS). In addition, when handling longer videos, \eg videos in the real world or from OVIS dataset, offline methods require splitting the input video into clips to avoid exceeding computational limits, and thus hand-designed clip matching is still inevitable, which will further decrease the overall performance. To sum up, matching/association is the main reasoning for the performance gap, and it is still inevitable and of great importance for offline models.

To improve the matching performance and thus bridge the performance gap, we propose a framework \textbf{I}n \textbf{D}efense of \textbf{O}n\textbf{L}ine models for video instance segmentation, termed IDOL. The key idea is to ensure, in the embedding space, the similarity of the same instance across frames and the difference of different instances in all frames, even for instances that belong to the same category and have similar appearances. It provides more discriminative instance features with better temporal consistency, which guarantees more accurate association results.
However, previous method~\cite{QDTrack} selects positive and negative samples by a hand-craft setting, introducing false positives in occlusions and crowded scenes, thus impairing contrastive learning. To address it, we formulate the problem of sample selection as an Optimal Transport problem in Optimization Theory, which reduces false positives and further improves the quality of the embedding.
During inference, by using one-to-many temporally weighted softmax, we utilize the learned prior of the embedding to re-identify missing instances caused by occlusions and to enforce the consistency and integrality of associations. 

Our thorough analysis gives us a deep understanding of current online and offline VIS methods. Based on our observation, we bridge the performance gap from the perspective of feature embeddings and propose IDOL. We conduct extensive experiments on YouTube-VIS 2019, YouTube-VIS 2021, and OVIS datasets. Despite its simplicity, our method sets a new state-of-the-art achieving 64.3 AP, 56.1 AP, and 42.6 AP on the validation set of these three benchmarks, respectively. More importantly, compared with previous online methods, we achieve a consistent improvement ranging from 13.2 AP to 14.7 AP on these datasets. We even surpass the previous SOTA offline method by up to 2.1 AP. We believe the simplicity and effectiveness of our method shall benefit further research. In addition, our thorough analysis provides insights for current methods and suggests useful directions for future work in both online and offline VIS.

\section{Related Work}

\noindent\textbf{Online Video Instance Segmentation.}
Most online VIS methods are built upon image-level instance segmentation with an additional tracking head to associate instances across the video. The baseline method MaskTrack R-CNN~\cite{MaskTrackRCNN} is built upon Mask R-CNN and proposes to leverage multi cues such as appearance similarity, semantic consistency, spatial correlation, and detection confidence to determine the instance labels. 
Most online methods~\cite{sipmask,compfeat,STMask,CrossVIS,STC} follow this pipeline. 
CrossVIS~\cite{CrossVIS} proposes a new learning scheme that uses the instance feature in the current frame to segment the same instance in another frame. 
Multi-Object Tracking and Segmentation (MOTS)~\cite{motchallenge,MOTS} aims to simultaneously segment and track all object instances of a given video sequence in real-time, which is similar to online VIS. MOTS methods are usually built upon multiple object trackers~\cite{Tracktor,QDTrack,FairMOT,CenterTrack,TransTrack,bytetrack,unicorn,PCAN}. Track R-CNN~\cite{MOTS} firstly extends the popular task of multi-object tracking to MOTS based on Mask R-CNN~\cite{MaskRCNN}. PointTrack~\cite{PointTrackV2} proposes a new online tracking-by-points paradigm with learning discriminative instance embeddings. Trackformer~\cite{Trackformer} adopts the transformer architecture for MOTS and introduces track query embeddings that follow objects through a video sequence.
Online models have a wider range of application scenarios, however, they are usually inferior to offline art by over 10 AP. We find that the current SOTA online models fail to achieve accurate associations,
causing the performance gap. We aim to tackle this problem in this work.


\noindent\textbf{Offline Video Instance Segmentation.}
Offline methods for VIS take the whole video as input and predict the instance sequence of the entire video (or video clip) with a single step. 
MaskProp~\cite{MaskProp} and Propose-Reduce~\cite{ProposeReduce} perform mask propagation in a video clip to improve mask and association. However, the propagation process is time-consuming, which limits its application.
Recently, VisTR~\cite{VisTR} adopts the transformer~\cite{transformers} to VIS and models the instance queries for the whole video. However, it learns an embedding for each instance of each frame, which makes it hard to apply to longer videos and more complex scenes.
IFC~\cite{IFC} proposes inter-frame communication transformers and significantly reduces computation and memory usage.
SeqFormer~\cite{seqformer} dynamically allocates spatial attention on each frame and learns a video-level instance embedding, which greatly improves the performance.
We deeply analyze the current SOTA offline models, IFC and SeqFormer, find that their improvement mainly comes from the black-box association between frames, 
but this advantage is gradually lost in complex scenarios.
{In contrast, our online method can be applied to both ongoing and long videos and complex scenarios, with more stable association quality and higher performance.}

\noindent\textbf{Contrastive learning} has made significant progress in representation learning~\cite{simclr,MOCO,QDTrack,coclr,predictivecoding,representation,non-parametric,Circleloss}. MOCO~\cite{MOCO} and SimCLR~\cite{simclr} use contrastive learning for image-level self-supervised training and learn strong feature representations for downstream tasks.
Some methods~\cite{coclr,supervised,videorep} extend the contrastive learning into multiple positive samples format to obtain better feature representations.
We absorb ideas from contrastive learning and propose to learn contrastive embeddings between frames for each instance.

\section{Method}

Given a video clip that consists of multiple image frames, online VIS models~\cite{MaskTrackRCNN,CrossVIS} utilize additional association head upon on instance segmentation models~\cite{MaskRCNN,CondInst}. 
We have already discussed that achieving more stable and discriminative instance embeddings between frames is the key to improve the performance of online models. To achieve this, we propose a contrastive learning framework to extract more discriminative features for instance association. 
We first introduce the instance segmentation pipeline in Sec.~\ref{sec:InstSeg}. Then the details of our contrastive learning framework and the cross-frame instance association strategy are introduced in Sec.~\ref{sec:Contrastive} and Sec.~\ref{sec:InstAssociation} respectively.

\subsection{Instance Segmentation.}
\label{sec:InstSeg}
For fair comparisons with the state-of-the-art offline method~\cite{seqformer}, we take DeformableDETR~\cite{deformableDETR} with dynamic mask head~\cite{CondInst} as our instance segmentation pipeline in this paper.
Our method can be coupled with other instance segmentation methods with minor modifications.

Given an input frame $x\in R^{3\times H\times W}$ of a video, a CNN backbone extracts multi-scale feature maps. The Deformable DETR module takes the feature maps with additional fixed positional encodings~\cite{detr} and $N$ learnable object queries as input. The object queries are first transformed into output embeddings $ E \in R^{N\times C} $ by the transformer decoder. After that, they are decoded into box coordinates and class labels by 3-layer feed-forward network (FFN) separately. 
For per-frame mask generation, we employ an FPN-like~\cite{FPN} mask branch to make the use of multi-scale feature maps from transformer encoder and generate feature map $F_{mask}$ that
are 1/8 resolution of the input frame.  
Another FFN encode outputs embeddings into parameters $\omega$ of mask head, which performs three-layer $1\times1$ convolution on the given feature map $F_{mask}$:
\begin{equation} \label{eq:dynamic_conv}
\textbf{m}_i = \text{MaskHead}(\textbf{F}_{\text{mask}},\omega_i).
\end{equation}
Then we calculate pair-wise matching cost which takes into account both the class prediction and the similarity of predicted and ground truth boxes.
For each ground truth, we assign multi predictions to it by selecting the top k predictions with the least cost by an optimal transport method~\cite{OTA,yolox}. Finally, the whole model is optimized with a multi-task loss function
\begin{equation} \label{eq:sumloss}
\mathcal{L} = \mathcal{L}_{cls}+\lambda_1\mathcal{L}_{box}+ \lambda_1\mathcal{L}_{mask}+\lambda_2\mathcal{L}_{embed},
\end{equation}
where loss weights $\lambda_1$ and $\lambda_2$ are set to 2.0 and 1.0 by default. For $\mathcal{L}_{box}$, we use a combination of $\mathcal{L}_{1}$ loss and the generalized IoU loss~\cite{giou}. The $\mathcal{L}_{mask}$ is defined as a combination of the Dice loss~\cite{diceloss} and Focal loss~\cite{focalloss}. $\mathcal{L}_{embed}$ is the contrastive loss described in the next section.

\begin{figure}[t]
	\centering
	\includegraphics[width=0.9\textwidth]{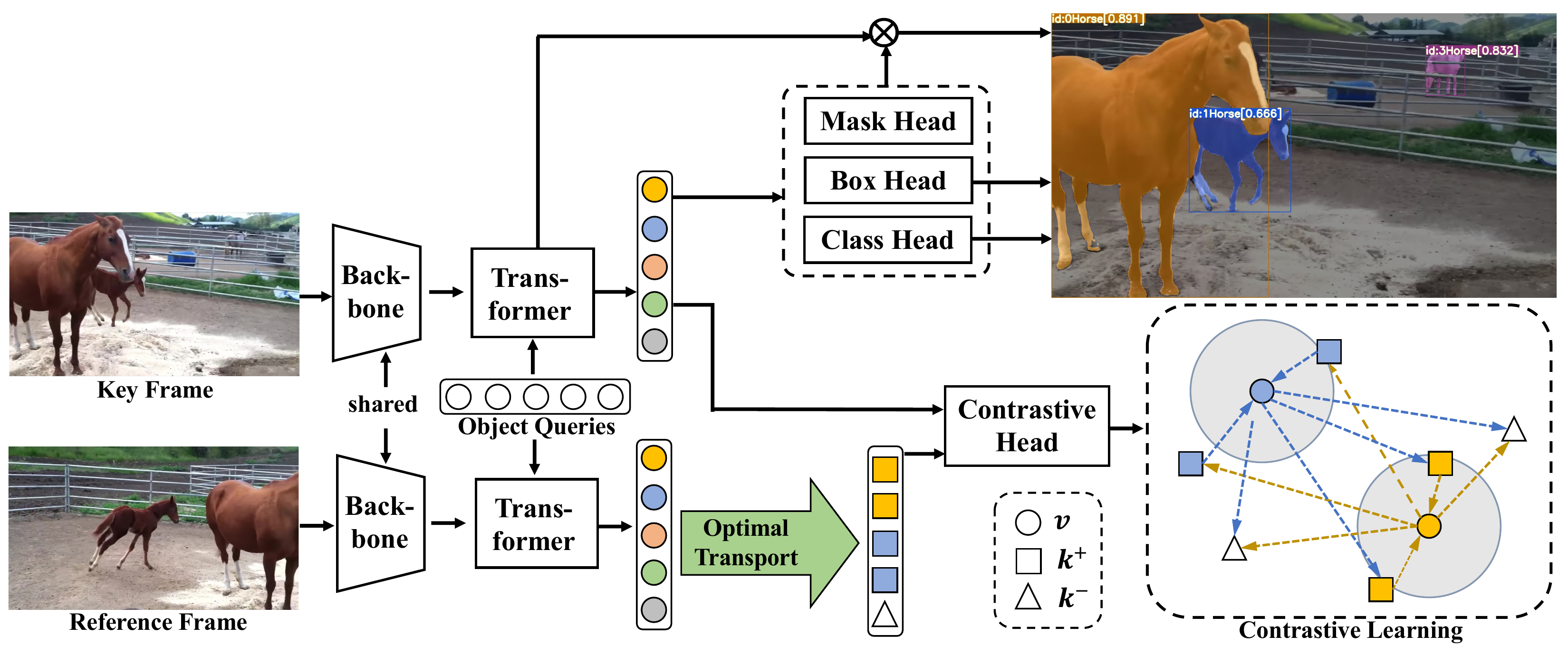}
	\caption{The training pipeline of IDOL. 
	Given a key frame and a reference frame, the shared-weight backbone and transformer predict the instance embeddings on them respectively. The embeddings on the key frame are used to predict masks, boxes, and categories, while the embeddings on the reference frame are selected as positive and negative embeddings dynamically by our optimal transport progress.
	Embeddings of the same color belong to the same video instance. Best viewed in color.}
	\label{fig:training}
\end{figure}

\subsection{Contrastive Learning between Frames.}
\label{sec:Contrastive}

More discriminative feature embeddings can help distinguish instances on different frames, thereby improving the quality of cross-frame association. To this end, we introduce contrastive learning between frames to make the embedding of the same object instance closer in the embedding space, and the embedding of different object instances farther away.
Object queries are used to query the features of instances from each frame in our instance segmentation pipeline. Therefore, the output embeddings can be regarded as features of different instances. We employ an extra light-weighted FFN as a contrastive head to decode the contrastive embeddings from the instance features.

Given a key frame for instance segmentation training, we select a reference frame from the temporal neighborhood. The instances appearing on the key frame may have different positions and appearances on the reference frame, but their contrastive embeddings should be as close as possible in embedding space.
For each instance in the key frame, we send the output embedding with the lowest cost to the contrastive head and get the contrastive embedding $\textbf{v}$. 
Different from previous method~\cite{QDTrack}, which selects positive and negative samples by a hand-craft setting, if the same instance appears on the reference frame, we take top $m1$ predictions with the least cost as positive and top $m2$ predictions with the highest cost as negatives. $m1$ and $m2$ are calculated dynamically by the optimal transport method~\cite{OTA,yolox}.
Please refer to the supplementary for more details.
The contrastive loss function for a positive pair of examples is defined as follows:
\begin{equation} \label{eq:lossembed1}
\begin{aligned}
\mathcal{L}_{embed}= &-\log \frac{\exp(\textbf{v} \cdot \textbf{k}^+)}{\exp(\textbf{v} \cdot \textbf{k}^+) + \sum_{\textbf{k}^-}\exp(\textbf{v}\cdot \textbf{k}^-)} \\
= & \log [1+\sum_{\textbf{k}^-}\exp(\textbf{v} \cdot \textbf{k}^-  - \textbf{v} \cdot \textbf{k}^+) ],
\end{aligned}
\end{equation}
where $\textbf{k}^+$ \text{and} $\textbf{k}^-$ are positive and negative feature embeddings from the reference frame, respectively.
We extend Eq.~\ref{eq:lossembed1} to multiple positive scenarios:
\begin{equation} \label{eq:lossembed2}
 \mathcal{L}_{embed} = \log [1+\sum_{\textbf{k}^+}\sum_{\textbf{k}^-}\exp(\textbf{v} \cdot \textbf{k}^-  - \textbf{v} \cdot \textbf{k}^+) ].
\end{equation}

\subsection{Instance Association.}
\label{sec:InstAssociation}
Previous online methods~\cite{MaskTrackRCNN,CrossVIS} take semantic consistency, spatial correlation, and detection confidence as cures. They are then leveraged to determine the instance labels. Other clip-based nearly online methods~\cite{MaskProp,STEmSEG,IFC} match instances using the predicted masks of overlapping frames by masking soft IoU metric between clips.
However, the online models perform instance segmentation on each frame independently, and therefore, the prediction quality on each frame is unstable. What makes it worse is the complex motion patterns, severe occlusions, false positives, duplicate predictions, error accumulation in long videos, and the frequently disappear and reappear objects, which makes instance association very challenging. Therefore, a strong instance association method should be robust to these cases. To this end, we propose a temporally weighted softmax score for instance matching and a memory bank-based association strategy to address these problems and improve the association quality of the online model.

\noindent\textbf{Temporally Weighted Softmax.}
Considering scenarios with fast motion, occlusion, and crowded objects, box-based matching introduces wrong association due to ambiguous location priors. However, the contrastive embedding learned by our method is able to maintain discriminative in embedding space when the position and shape change.
Therefore, the contrastive embeddings are used to calculate the embedding similarity between the instances on the current and the previous frames.
Assume there are $N$ predicted instances with $N$ contrastive embeddings ${\textbf{d}_i} \in \mathbb{R}^{C} $, and $M$ instances in the memory bank, each of which has multiple temporal contrastive embeddings $\{\textbf{e}_j^t\}^T_{t=1}, e_j^t\in \mathbb{R}^{C}$ from previous $T$ frames.
These embeddings are combined by a temporally weighted sum:
\begin{equation} \label{eq:weightedsum}
\hat{\textbf{e}_j} = \frac{\sum_{t=1}^T \textbf{e}_j^t \times (\tau+T/t )   }{\sum_{t=1}^T \tau+T/t}.
\end{equation}
Then we compute bi-directional similarity $f$ between predicted instance $i$ and memory instance $j$ by:
\begin{equation} \label{eq:frameweight}
\textbf{f}(i,j) = [\frac{\exp(\hat{\textbf{e}_j} \cdot \textbf{d}_i)+\sigma_j}{\sum^M_{k=1} \exp(\hat{\textbf{e}_k} \cdot \textbf{d}_i) + \sigma_k   } + \frac{\exp(\hat{\textbf{e}_j} \cdot \textbf{d}_i)}{\sum^N_{k=1} \exp(\hat{\textbf{e}_j} \cdot \textbf{d}_k)   }]/2,
\end{equation}
where $\sigma_j$ is the existing time of instance $j$ in the memory, it serves as the confidence scores of each instance in the memory.
By introducing the temporal contrastive embeddings and the confidence scores determined by the duration of existence, 
the learned prior information is able to reidentify missing instances caused by occlusions, enforcing the consistency and integrality of associations.

\noindent\textbf{Association Strategy.}
To take full advantage of the learned contrastive embedding, we propose a new association strategy during inference.
Given a test video, we initialize an empty memory bank for it and perform instance segmentation on each frame sequentially in an online scheme. For the prediction of each frame, we first perform inter-class duplicate removal by NMS with a threshold of 0.5.
Then we compute matching scores $\textbf{f}(i,j)$ between predictions and memory bank by Eq.~\ref{eq:frameweight}, and search for the best assignment for instance $i$ by: 
\begin{equation} \label{eq:argmax}
\hat{j} = \arg\max \textbf{f}(i,j), \forall j \in\{1,2,...,M\}.
\end{equation}
If $\textbf{f}(i,\hat{j})>0.5$, we assign the instance $i$ on current frame to the memory instance $\hat{j}$. For the prediction without an assignment but has a high class score, we start a new instance ID in the memory bank.  

\section{Experiments}

\subsection{Dataset and Metrics}
\label{sec:dataset}
We report our results on YouTube-VIS 2019~\cite{MaskTrackRCNN}, YouTube-VIS 2021~\cite{ytvis21dataset}, and OVIS~\cite{ovis} datasets. YouTube-VIS 2019 is the first and largest dataset for video instance segmentation. It contains 2,238 training, 302 validation, and 343 test high-resolution YouTube video clips, with an average video duration of 4.61s. 
YouTube-VIS 2021 is an extended version of YouTube-VIS 2019. 
Both datasets have 40  object categories, but the category label set is slightly different. 
OVIS dataset is a relatively new and challenging dataset. It consists of 607 training videos, 140 validation videos, and 154 test videos. Compared with YouTube-VIS, its videos are much longer and last 12.77s on average, and more importantly, it contains much more videos that record objects with severe occlusion, complex motion patterns, and rapid deformation. 
All these features make OVIS an ideal dataset to evaluate and analyze different methods.
We report standard metrics such as $\text{AP}, \text{AP}_{50}, \text{AP}_{75}, \text{AR}_{1}, \text{and}\ \text{AR}_{10}$. IoU threshold is used during evaluation.

\subsection{Implementation Details}
\noindent\textbf{Model settings.} We use ResNet-50~\cite{resnet} as our backbone unless otherwise specified. For a fair comparison with SOTA offline method, we use the same setting for Deformable DETR and the dynamic mask head following SeqFormer~\cite{seqformer}. For the transformer, we use 6 encoders, 6 decoder layers of width 256 with bounding box refinement mechanism, and the number of object queries is set to 300.

\noindent\textbf{Training.}
We use AdamW~\cite{AdamW} optimizer with base learning rate of $1\times 10^{-4}$, and weight decay of $10^{-4}$. We first pre-train the model on COCO for instance segmentation following previous work~\cite{VisTR,IFC,CrossVIS,seqformer}. 
Then we train our model for 12000 iterations, on the corresponding training set and reduce learning rate by a factor of 10 at at the 8000 iterations.
For the result with superscript ``\dag",
we randomly and independently crop the image from COCO twice to form a pseudo key-reference frame pair, which is used to pre-train the contrastive embedding of our models before training on video datasets.
For YouTube-VIS 2019 and YouTube-VIS 2021, the input frames are downsampled and randomly cropped so that the longest side is at most 768 pixels. For OVIS, we use the same scale augmentation with COCO, resizing the input images so that the shortest
side is at least 480 and at most 800 pixels while the longest is at most 1333. The model is trained on 8 V100 GPUs of 32G RAM, with 2 pairs of frames per GPU.
In order to provide more positive embeddings for contrastive training, we adopt optimal transport theory~\cite{OTA,yolox} to assign a ground truth label to multiple predictions. 
Please refer to the supplementary for more details about OT.

\begin{sloppypar} 
\noindent\textbf{Inference.}
During inference, the input frames are downscaled to 360p for YouTube-VIS 2019 and YouTube-VIS 2021 following previous work, and 720p for OVIS as its videos has a higher resolution. For the  hyper-parameters of temporally weighted softmax, we set $\tau=0.5$ and $\text{T}=3$ by default. 
\end{sloppypar}

\subsection{Analysis of Current SOTA VIS Models}
\label{sec:VIS-Analysis}
Since no annotation for the validation set is available, we split the original training set into custom training split and validation split. All models are trained on the training split and evaluated on the validation split. YouTube-VIS 2019 and OVIS are used. We analyze the results as follows:

\noindent\textbf{Performance gaps between online and offline models: }
First, we compare the mask quality of two recent online and offline methods in Table.~\ref{table:crossvis_IFC}. 
They are both published in 2021 thus can be considered as work in the same period.
When ground-truth instance ID is provided (the column of `frame oracle'), CrossVIS outperforms IFC on both YouTube-VIS and OVIS. However, when the instance ID is not provided (the column of `predicted'), the methods are required to match the results, and the performance of CrossVIS drops dramatically by 9.4 AP while the performance of IFC drops by 3.3 AP on YouTube-VIS, leading to the poor performance of CrossVIS. Offline methods enable the model to match predicted masks by itself and avoid using hand-designed association modules. It works well on simple datasets and benefits current offline models, but it still fails on 
challenging datasets like OVIS.

\setlength{\tabcolsep}{4pt}
\begin{table}[t]
\begin{center}
\caption{Oracle experiments on association quality. Frame oracle means gt instance ID is provided (same to Fig.~\ref{fig:motivation}). We set the clip length of IFC to 10. AP is reported.}
\label{table:crossvis_IFC}
\resizebox{0.7\columnwidth}{!}{
\begin{tabular}{lllcc}
\hline\noalign{\smallskip}
Dataset   &Method & Publish &Predicted  &Frame Oracle  \\
\noalign{\smallskip}
\hline
\noalign{\smallskip}
\multirow{2}{*}{YouTube-VIS}
 &CrossVIS & ICCV 2021 &43.4 &52.8  \\
 &IFC & NeurIPS 2021 &46.8 &50.1 \\
\hline
\multirow{2}{*}{OVIS}
 &CrossVIS & ICCV 2021 &10.1 &29.9  \\
 &IFC & NeurIPS 2021 &8.7 &25.1 \\
\hline
\end{tabular}}
\end{center}
\vspace{-2em}
\end{table}
\setlength{\tabcolsep}{1.4pt}

\setlength{\tabcolsep}{4pt}
\begin{table}[t]
\begin{center}
\caption{Oracle experiments on clip length for offline models. AP is reported.}
\label{table:offline_model_oracle}
\resizebox{0.75\columnwidth}{!}{
\begin{tabular}{lcccccccc}
\hline\noalign{\smallskip}
\multirow{2}{*}{Dataset}   &\multirow{2}{*}{Method}  &\multirow{2}{*}{Oracle Type} &\multicolumn{6}{c}{Clip Length (frame)} \\
 \cmidrule(lr){4-9}
 & & &1 &3 &5 &10 &20 &30 \\
\noalign{\smallskip}
\hline
\noalign{\smallskip}
\multirow{4}{*}{YouTube-VIS}
 &\multirow{2}{*}{IFC} &frame &48.1 &49.5 &49.7 &50.1 &51.3 &51.8  \\
 & &clip &48.1 &49.1 &49.0 &48.9 &49.7 &50.1  \\
 \cmidrule(lr){2-9}
 &\multirow{2}{*}{SeqFormer} &frame &57.9 &58.2 &57.9 &57.4 &57.6 &57.0 \\
 &  &clip &57.9 &53.8 &53.14 &52.4 &50.2 &50.2 \\
\hline
\multirow{4}{*}{OVIS}
 &\multirow{2}{*}{IFC} &frame &24.6 &25.8 &25.9 &25.1 &24.2 &22.8  \\
 & &clip  &24.6 &23.6 &22.5 &18.8 &13.5 &10.5  \\
 \cmidrule(lr){2-9}
 &\multirow{2}{*}{SeqFormer} &frame &32.0 &32.3 &31.8 &31.5 &30.3 &29.8 \\
 &  &clip &32.0 &28.3 &25.0 &18.9 &11.5 &8.9 \\
\hline
\end{tabular}}
\end{center}
\vspace{-2em}
\end{table}
\setlength{\tabcolsep}{1.4pt}

\begin{figure}[t]
	\centering
	\includegraphics[width=0.8\textwidth]{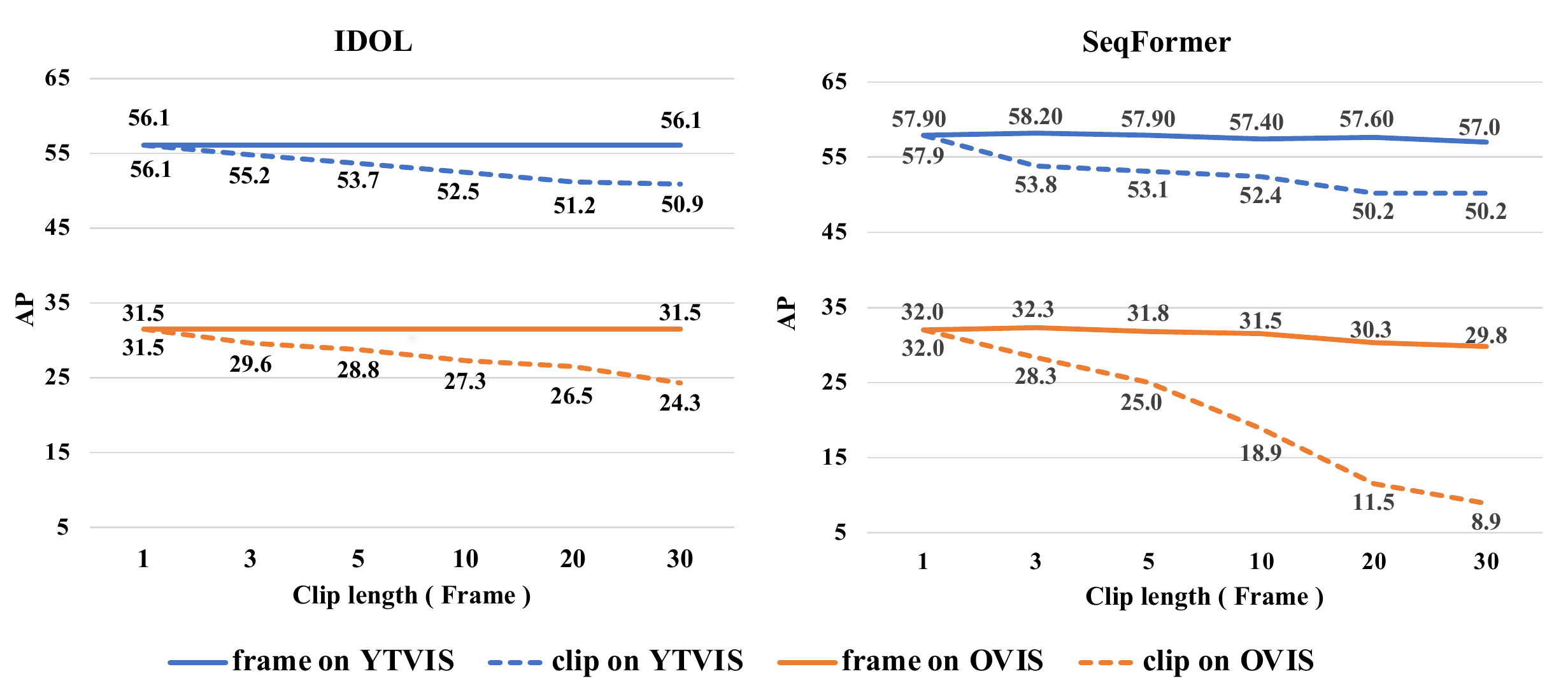}
	\caption{\textbf{Oracle experiments on IDOL and SeqFormer. }}
	\label{fig:idol_oracle}
\end{figure}

\begin{figure*}[t]
\centering
\includegraphics[width=1\linewidth]{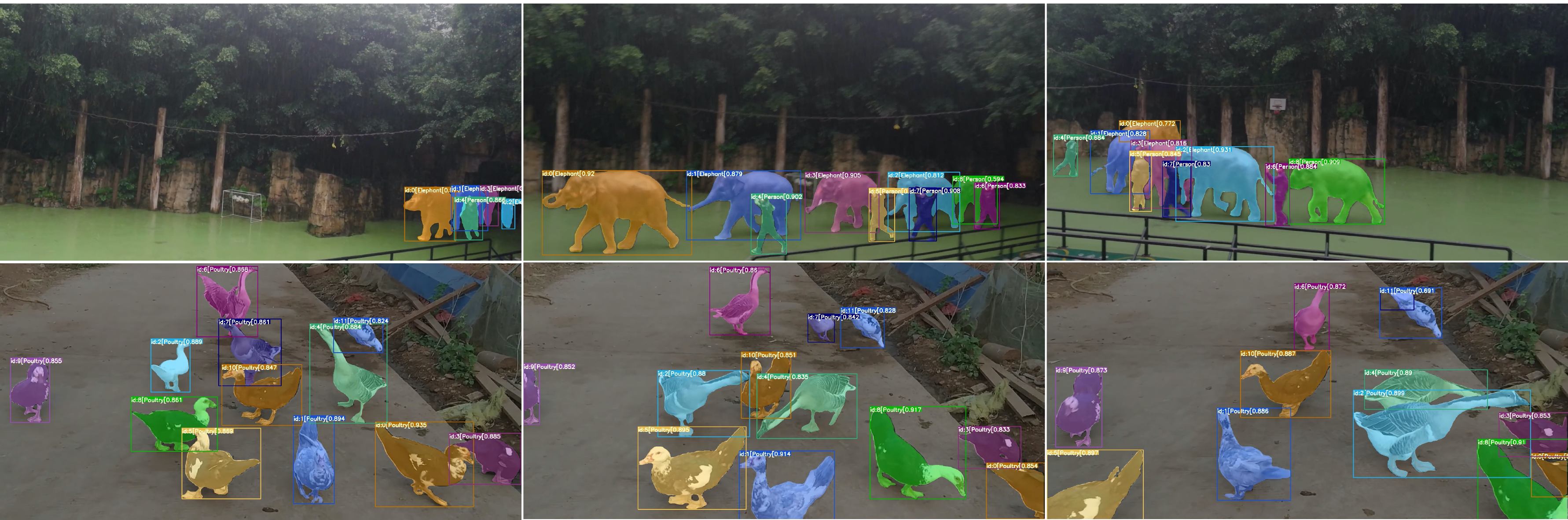}
\caption{Qualitative results on the OVIS validation dataset. IDOL achieves very good results on complex scenes. Please refer to the supplementary for more qualitative results and comparison with other methods.}
\label{fig:ovis_visualization}
\end{figure*}

\noindent\textbf{Analysis of current offline models: }
In Table.~\ref{table:offline_model_oracle}, we give detailed analyses for SOTA offline models, hoping to provide insights for future research. Compared with online methods, offline models in theory have two inherent advantages:

First, as we mentioned above, it avoids hand-designed association. However, this step is very sensitive to the occlusion and the complexity of videos, especially when the clip becomes longer. For example, when clip length is equal to 5, the black-box association degrades the performance of IFC and SeqFormer on OVIS by 3.4 AP (25.9 AP \vs 22.5 AP) and 6.8 AP (31.8 AP \vs 25.0 AP), respectively. When clip length is set to 30, the performance drops by 12.3 AP (22.8 AP \vs 10.5 AP) and 20.9 AP (29.8 AP \vs 8.9 AP), respectively. What's more, even a clip length of 30 still doesn't meet the requirement of real-world application. Clip matching is still inevitable, and it further decreases the overall performance. 

Another inherent advantage of offline models is the ability to use multiple frames for instance segmentation, which provides more information to handle occlusion and optimize the results. However, current models still fail to fully utilize this feature. Currently, it only works for simple videos: compared with per-frame segmentation (clip length=1), it improves the mask quality by at most 3.7 AP for IFC (when clip length=30) and 0.3 AP for SeqFormer (when clip length=3). When testing on OVIS, multiple frames segmentation only improves the mask quality by at most 1.3 AP for IFC (clip length=5) and 0.3 AP for SeqFormer (clip length=3), and even degrades the performance by 1.8 AP for IFC and 2.2 AP for SeqFormer when clip size becomes longer (clip length=30). What's more, the improvement is even less obvious in practice when no ground-truth instance ID is provided, even for simple videos, due to the association problem. It only improves the performance of IFC on YouTube-VIS by 2.0 AP, but degrades the performance in all the other experiments.

To prove the effectiveness of our method, we further analyze IDOL with oracle experiments in Fig.\ref{fig:idol_oracle}.
Since IDOL is an online model, the results of frame oracles with different clip lengths are the same as per-frame segmentation oracle results.
For clip oracles, IDOL is required to do association within the clips by itself.
Compared with SeqFormer, the gaps between frame oracles and clip oracles of IDOL are much smaller on OVIS, proving that IDOL performs a more robust association between frames than the offline model on challenging datasets.

\subsection{Main Results}
We compare IDOL against current online and offline SOTA methods on YouTube-VIS 2019, YouTube-VIS 2021, and OVIS validation sets. The results are reported in Tables~\ref{table:mainYTVIS2019},~\ref{table:mainYTVIS2021}, and~\ref{table:mainOVIS}, respectively. We compare the methods with different backbones~\cite{resnet,SwinTransformer} for a fair comparison. Notably, our method significantly surpasses all previous online methods by at least $10.1$ AP. In addition, we also outperform all previous offline methods under all evaluation metrics when training on the same data. 
More importantly, our method achieves an overall first place~\cite{wu1st} in the YouTube-VIS Challenge 2022, which proves our SOTA performance.
In addition, our method only decreases the inference speed of the adopted instance segmentation pipeline by 1.1 FPS on an RTX-2080Ti, which proves our efficiency. In general, our method is simple and very effective compared with all baseline methods.
Qualitative results on sample videos of the challenging OVIS dataset are shown in Fig.~\ref{fig:ovis_visualization}. More qualitative results can be found in the supplementary.
We analyze the performance in detail as follows:

\setlength{\tabcolsep}{4pt}
\begin{table}[t]
\begin{center}
\caption{Comparison on YouTube-VIS 2019 val set. The best results with the same backbone are in \textbf{bold} and second \underline{underline}. `V' means only YouTube-VIS training set is used. `V+I' means synthetic videos from COCO with overlapping categories are also used for joint training. Note that offline models take advantage of larger batch sizes thus having much bigger FPS, while online models handle video frame by frame.
The result with superscript ``\dag" is obtained by pre-training on COCO pseudo key-reference frame pairs, and resolution of 480p is used during inference.
}
\label{table:mainYTVIS2019}
\resizebox{0.9\columnwidth}{!}{
\begin{tabular}{llcccccccc}
\hline\noalign{\smallskip}
Backbone   &Method  &Type &FPS &Data   &$\rm AP$    &$\rm AP_{50}$  &$\rm AP_{75}$ &$\rm AR_{1}$  &$\rm AR_{10}$  \\
\noalign{\smallskip}
\hline
\noalign{\smallskip}
\multirow{14}{*}{ResNet-50}
&MaskTrack R-CNN~\cite{MaskTrackRCNN}   &{online} &20.0 &V  &30.3 &51.1 &32.6 &31.0 &35.5  \\ 
 &SipMask~\cite{sipmask}    &{online} &30.0 &V  &33.7 &54.1 &35.8 &35.4 &40.1   \\
 &CompFeat~\cite{compfeat}   &{online} &- &V  &35.3 &56.0 &38.6 &33.1 &40.3   \\
  &CrossVIS~\cite{CrossVIS}       &{online} &39.8 &V &36.3 &56.8 &38.9 &35.6 &40.7   \\
 &PCAN~\cite{PCAN}       &{online} &- &V &36.1 &54.9 &39.4 &36.3 &41.6   \\
  &STEm-Seg~\cite{STEmSEG}     &\gray{offline} &7.0  &V+I  &30.6 &50.7 &33.5 &37.6 &37.1 \\
 &VisTR~\cite{VisTR}         &\gray{offline} &69.9 &V  &36.2 &59.8  &36.9 &37.2 &42.4    \\
 
 &MaskProp~\cite{MaskProp}      &\gray{offline} &- &V  &40.0 &\ \ \ -    &42.9 &\ \ \   -   &\ \ \  -    \\  
 &Propose-Reduce~\cite{ProposeReduce}   &\gray{offline} &- &V+I  &40.4 &63.0 &43.8 &41.1 &49.7   \\
 &IFC~\cite{IFC}     &\gray{offline} &107.1 &V  &42.8 &65.8 &46.8 &43.8 &51.2  \\  
  &{SeqFormer}~\cite{seqformer}   &\gray{offline} &72.3  &V  &45.1 &66.9 &50.5 &\underline{45.6} &54.6  \\
&{SeqFormer}~\cite{seqformer}       &\gray{offline} &72.3  &V+I   &\underline{47.4} &{69.8} &{51.8} &{45.5} &{54.8}   \\
  &\textbf{IDOL(ours)}    &{online} &30.6 &V   &{46.4} &\underline{70.7} &\underline{51.9} &{44.8} &\underline{54.9} \\  
  
 &\textbf{IDOL(ours)$^{\dag}$}    &{online} &30.6 &V   &\textbf{49.5} &\textbf{74.0} &\textbf{52.9} &\textbf{47.7} &\textbf{58.7} \\  
 \midrule
 \multirow{11}{*}{ResNet-101}
 &MaskTrack R-CNN~\cite{MaskTrackRCNN}     &{online} &- &V  &31.8 &53.0 &33.6 &33.2 &37.6  \\
  &CrossVIS~\cite{CrossVIS}      &{online} &35.6 &V  &36.6 &57.3 &39.7 &36.0 &42.0 \\
 &PCAN~\cite{PCAN}  &{online} &- &V  &37.6 &57.2 &41.3 &37.2 &43.9   \\
 &STEm-Seg~\cite{STEmSEG}     &\gray{offline} &- &V+I  &34.6 &55.8 &37.9 &34.4 &41.6 \\
 &VisTR~\cite{VisTR}     &\gray{offline} &57.7 &V   &40.1 &64.0 &45.0 &38.3 &44.9  \\
&MaskProp~\cite{MaskProp}     &\gray{offline} &- &V   &42.5 &\ \ \  -   &45.6 &\ \ \  -   &\ \ \  -   \\  
 &Propose-Reduce~\cite{ProposeReduce}     &\gray{offline} &- &V+I  &43.8 &65.5 &47.4 &43.0 &53.2  \\
 &IFC~\cite{IFC}   &\gray{offline} &89.4 &V  &44.6 &69.2 &49.5 &44.0 &52.1  \\  
 &{SeqFormer}~\cite{seqformer}  & \gray{offline} &64.6 &V+I   &\underline{49.0} &{71.1} &\underline{55.7}   &\underline{46.8}  &\underline{56.9}  \\  
 &\textbf{IDOL(ours)}    &online &26.0 &V  &{48.2} &\textbf{73.6} &{52.5} &{45.6} &{55.5}  \\  
  &\textbf{IDOL(ours)$^{\dag}$}    &online &26.0 &V  &\textbf{50.1} &\underline{73.1} &\textbf{56.1} &\textbf{47.0} &\textbf{57.9}  \\  
 \midrule
 \multirow{3}{*}{Swin-L}
  &SeqFormer~\cite{seqformer}     &\gray{offline} &27.7 &V+I   &{59.3}  &{82.1}  &{66.4}  &{51.7}  &64.4 \\ 
   &\textbf{IDOL(ours)}    &{online} &17.6 &V   &\underline{61.5} &\underline{84.2} &\underline{69.3} &\underline{53.3} &\underline{65.6} \\  
   &\textbf{IDOL(ours)$^{\dag}$}    &{online} &17.6 &V   &\textbf{64.3} &\textbf{87.5} &\textbf{71.0} &\textbf{55.6} &\textbf{69.1} \\  
\hline
\end{tabular}
}
\end{center}
\vspace{-2em}
\end{table}

\noindent\textbf{YouTube-VIS 2019: } It is the most commonly used dataset. We reported the results in Table~\ref{table:mainYTVIS2019}. When using the same backbones, IDOL significantly outperforms previous online methods by 10.1 AP (on ResNet-50) and 10.6 AP (on ResNet-101). 
Compared with offline methods, we achieve better performance when no extra data is used for training, surpassing previous methods by 1.3 AP and 2.2 AP with ResNet-50 and Swin-L as the backbone, respectively. 
For the result with superscript ``\dag",
we randomly crop images from COCO twice to form pseudo key-reference frame pairs to pre-train the contrastive embedding part of our model before training on real video datasets.
It improves the performance by 3.1 AP on ResNet-50 and 2.8 AP on Swin-L backbone, outperforming previous SOTA models comprehensively.

\setlength{\tabcolsep}{4pt}
\begin{table}[t]
\begin{center}
\caption{Comparison on YouTube-VIS 2021 val set. Best in \textbf{bold}, second \underline{underline}.}
\label{table:mainYTVIS2021}
\resizebox{0.80\columnwidth}{!}{\begin{tabular}{llccccccc}
\hline\noalign{\smallskip}
Backbone   &Method  &Type  &$\rm AP$    &$\rm AP_{50}$  &$\rm AP_{75}$ &$\rm AR_{1}$  &$\rm AR_{10}$  \\
\noalign{\smallskip}
\hline
\noalign{\smallskip}
\multirow{7}{*}{ResNet-50}
 &MaskTrack R-CNN~\cite{MaskTrackRCNN}   &online   &28.6 &48.9 &29.6 &26.5 &33.8  \\
 &SipMask~\cite{sipmask}    &online    &31.7 &52.5 &34.0 &30.8 &37.8    \\
 &STMask~\cite{STMask}   &online   &31.1 &50.4 &33.5 &26.9 &35.6  \\
  &CrossVIS~\cite{CrossVIS}       &online   &34.2 &54.4 &37.9 &30.4 &38.2   \\
  &IFC~\cite{IFC}    & \gray{offline} &36.6 &57.9 &39.3 &- &- \\
  &SeqFormer~\cite{seqformer}  &\gray{offline} &\underline{40.5} &\underline{62.4} &\underline{43.7} &\underline{36.1} &\underline{48.1}  \\
  &\textbf{IDOL(ours)}    &online   &\textbf{43.9} &\textbf{68.0} &\textbf{49.6} &\textbf{38.0} &\textbf{50.9} \\  
\hline
\multirow{2}{*}{Swin-L }
    &{SeqFormer}~\cite{seqformer} &\gray{offline} &51.8 &74.6 &58.2 &42.8   &58.1 \\
   &\textbf{IDOL(ours)}    &online  &\textbf{56.1} &\textbf{80.8} &\textbf{63.5} &\textbf{45.0} &\textbf{60.1}  \\  
\hline
\end{tabular}}
\end{center}
\end{table}
\setlength{\tabcolsep}{1.4pt}

\setlength{\tabcolsep}{4pt}
\begin{table}[tb]
\begin{center}
\caption{Comparison on OVIS 2021 val set. Best in \textbf{bold} and second \underline{underline}. The results with superscript ``\dag" are not reported in~\cite{IFC,seqformer}. Videos in OVIS are much longer than those in YTVIS, offline models are unable to take the entire video as input due to the limit of computational resources. We split the video into clips of length 10 and perform clip matching provided by~\cite{IFC} on overlapping frames to get the final results.}
\label{table:mainOVIS}
\resizebox{0.82\columnwidth}{!}{\begin{tabular}{llclllll}
\hline\noalign{\smallskip}
Backbone   &Method  &Type &$\rm AP$    &$\rm AP_{50}$  &$\rm AP_{75}$ &$\rm AR_{1}$  &$\rm AR_{10}$  \\
\noalign{\smallskip}
\hline
\noalign{\smallskip}
\multirow{8}{*}{ResNet-50}
 &MaskTrack R-CNN~\cite{MaskTrackRCNN}   &{online}  &10.8 &25.3 &8.5 &7.9 &14.9  \\
 &SipMask~\cite{sipmask}    &{online}   &10.2 &24.7 &7.8 &7.9 &15.8   \\
 &CMaskTrack R-CNN~\cite{ovis}   &{online}  &\underline{15.4} &\underline{33.9} &13.1 &9.3 &20.0  \\
  &CrossVIS~\cite{CrossVIS}       &{online}  &14.9 &32.7 &12.1 &10.3 &19.8   \\
  &STEm-Seg~\cite{STEmSEG}     &\gray{offline}  &13.8 &32.1 &11.9 &9.1 &20.0 \\
  &{IFC}$^{\dag}$~\cite{IFC}  &\gray{offline}  &13.1 &27.8 &11.6 &9.4 & 23.9 \\
  &{SeqFormer}$^{\dag}$~\cite{seqformer} &\gray{offline} &15.1 &31.9 &\underline{13.8} &\underline{10.4} &\underline{27.1} \\
  &\textbf{IDOL(ours)}    &{online}   &\textbf{30.2} &\textbf{51.3} &\textbf{30.0} &\textbf{15.0} &\textbf{37.5}  \\  
 \midrule
 Swin-L   &\textbf{IDOL(ours)}    &{online}   &\textbf{42.6} &\textbf{65.7} &\textbf{45.2} &\textbf{17.9} &\textbf{49.6}  \\  
\hline
\end{tabular}}
\end{center}
\end{table}

\noindent\textbf{YouTube-VIS 2021: } It is an extended version of YouTube-VIS 2019. It contains more videos with a larger number of instances and frames. As shown in Table~\ref{table:mainYTVIS2021}, we achieve 43.9 AP with a ResNet-50 backbone, surpassing the previous best online method and offline method by 9.7 AP and 3.4 AP, respectively.

\setlength{\tabcolsep}{1.4pt}

\noindent\textbf{OVIS: } As mentioned before, OVIS contains long video sequences with heavy occlusion and complex motion, thus it is extremely difficult for all algorithms and exceeds the capability of offline methods due to the limit of computational resources. STEm-Seg~\cite{STEmSEG} is the only offline method that can be directly evaluated on OVIS since its design enables it to run in a nearly online manner. To compare with the SOTA offline methods (\eg IFC~\cite{IFC} and SeqFormer~\cite{seqformer}), we split the video into short clips and apply the clipping matching method provided in IFC on these two methods (SeqFormer doesn't provide its matching method). The results are provided in Table~\ref{table:mainOVIS}. Note that the previous best method only gains 15.4 AP on the validation set. IDOL with the same ResNet-50 backbone achieves $2\times$ performance and gains 30.2 AP, surpassing the previous method by 14.8 AP. What's more, when using a stronger backbone (\ie, Swin-L) to extract better features, IDOL achieves the state-of-the-art performance of 42.6 AP, which is a huge improvement over previous best results.

\setlength{\tabcolsep}{4pt}
\begin{table}[t]
\begin{center}
\caption{Ablation study on contrastive learning and inference strategies on YTVIS.}
\label{table:trainval_ablation_ytvis}
\resizebox{0.70\columnwidth}{!}{\begin{tabular}{cccccccc}
\hline\noalign{\smallskip}
\multicolumn{3}{c}{Training} &\multicolumn{2}{c}{Inference} &\multirow{2}{*}{$\rm AP$} &\multirow{2}{*}{$\rm AP_{75}$} &\multirow{2}{*}{$\rm AR_{1}$}\\
\cmidrule(lr){1-3} \cmidrule(lr){4-5}
ID Head &Contrastive  &OT  &Matching &Temporal \\ 
\noalign{\smallskip}
\hline
\noalign{\smallskip}
\checkmark &- &-  &multi-cues &- &30.3 &30.5 &31.5  \\
\checkmark &- &-  &embeddings &- &33.6 &36.8 &38.7 \\
\checkmark &- &-  &embeddings &\checkmark &34.8 &37.6 &39.4 \\
- &\checkmark &-  &multi-cues &- &31.8 &35.3 &31.9 \\
- &\checkmark &-  &embeddings &- &42.5 &45.7 &42.2 \\
- &\checkmark &\checkmark   &embeddings &- &44.8 &49.9 &43.4 \\
- &\checkmark &\checkmark  &embeddings &\checkmark &46.4 &51.9 &44.8 \\
\hline
\end{tabular}}
\end{center}
\end{table}
\setlength{\tabcolsep}{1.4pt}

\setlength{\tabcolsep}{4pt}
\begin{table}[t]
\begin{center}
\caption{Ablation study on contrastive learning and inference strategy on OVIS. Medium and heavy denote the AP and AR of objects moderately occluded, and heavily occluded, respectively.}
\label{table:trainval_ablation_ovis}
\resizebox{0.9\columnwidth}{!}{\begin{tabular}{cccccccccc}
\hline\noalign{\smallskip}
\multicolumn{3}{c}{Training} &\multicolumn{2}{c}{Inference} &\multicolumn{3}{c}{$\rm AP$} &\multicolumn{2}{c}{$\rm AR$}  \\
\cmidrule(lr){1-3} \cmidrule(lr){4-5} \cmidrule(lr){6-8} \cmidrule(lr){9-10}
ID Head &Contrastive  &OT  &Matching &Temporal &All &medium &heavy &medium &heavy  \\ 
\noalign{\smallskip}
\hline
\noalign{\smallskip}
\checkmark &- &-  &multi-cues &- &11.0 &11.9 &2.3 &16.4 &7.6  \\
- &\checkmark &-  &multi-cues &- &18.4 &22.5 &5.8 &34.9 &14.9 \\
- &\checkmark &-  &embeddings &- &26.7 &30.9 &9.5 &43.2 &19.9 \\
- &\checkmark &\checkmark   &embeddings &- &28.3 &34.0 &9.8 &44.5 &20.3 \\
- &\checkmark &\checkmark  &embeddings &\checkmark &30.2 &36.5 &10.3 &46.9 &20.5 \\
\hline
\end{tabular}}
\end{center}
\vspace{-2em}
\end{table}
\setlength{\tabcolsep}{1.4pt}

\subsection{Ablation Study}

In this section, we conduct extensive ablation experiments to study the importance of the core factors of our method on YouTube-VIS 2019 and OVIS.
Previous SOTA online method~\cite{CrossVIS} uses an extra M-class classification head where $M$ equals to the number of all instances in the training set, termed as ``ID Head'' in Table~\ref{table:trainval_ablation_ytvis}. and~\ref{table:trainval_ablation_ovis}. 
In the ``Contrastive" setting, we use box IoU between predictions and ground truth for positive and negative embeddings selection following~\cite{QDTrack}. We further evaluate our optimal transport method to dynamically select positive and negative embeddings, termed as ``OT".
For ablation study on inference strategy, ``multi-cues'' setting combines semantic consistency, spatial correlation, detection confidence and appearance similarity together to perform association following~\cite{MaskTrackRCNN,CrossVIS}. Our association strategy is termed ``embedding".

\noindent\textbf{Contrastive Training.}
To evaluate the importance of our contrastive embeddings, we apply the same association method on the embeddings predicted by ID Head and contrastive head. As shown in Table~\ref{table:trainval_ablation_ytvis}, contrastive training only improves 1.5 AP with the multi-cues association but improves 8.9 AP when it comes to the embedding-based association.
Our explanation is that contrastive training provides more discriminative embeddings for instance association, but other cues in multi-cues weaken the role of embeddings. 
In addition, on the more challenging OVIS dataset, contrastive embeddings increase AP from 11.0 to 18.4, which brings an improvement of 67.3\%.
This indicates that embedding-based association is more robust in longer videos and complex scenarios.
Furthermore, optimal transport matching (OT) improves the results by 2.3 AP on YTVIS and 2.3 AP on OVIS, which indicates that the choice of positive and negative embeddings plays an important role in learning discriminative embeddings. OT provides a better selection of positive and negative embeddings during training, improving the quality of embeddings. We show the visualization of positive and negative embeddings selected by these two strategies in supplementary.

\noindent\textbf{Association Strategy and Temporally Weighted Softmax.}
As shown in Table~\ref{table:trainval_ablation_ytvis}, compared with ``multi-cues", our embedding association strategy takes advantage of the discriminative embedding learned by contrastive learning, and improves the AP from 31.8 to 42.5 on YouTube-VIS. 
When it comes to OVIS in Table~\ref{table:trainval_ablation_ovis}, our association strategy also improves the AP from 18.4 to 26.7.
In addition, when temporally weighted softmax is added, it can be further improved by 1.6 AP on YouTube-VIS and 1.9 AP on OVIS. Utilizing information and priors from multiple previous frames improve robustness of association. Considering the problem of false positives and disappearing-and-reappearing that the online model needs to deal with, we believe this strategy helps maintain temporal consistency. We provide more visualization results, detailed analysis, and additional ablation experiments on OVIS in supplementary.

\section{Conclusions}

Online video instance segmentation methods have their inherent advantage in handling long/ongoing videos, but they are inferior to the offline models in performance. In this work, we aim to bridge the performance gap. We first deeply analyze the current online and offline models and find that the gap mainly comes from the error-prone association between frames. Based on this observation, we propose IDOL, which enables models to learn more discriminative and robust instance features for VIS tasks. It significantly outperforms all online and offline methods and achieves new SOTA on three benchmarks. We believe our insights on VIS methods will inspire future work in both online and offline methods.

~\\
\noindent{\textbf{Acknowledgements }}
This work was supported by NSF 1763705.
We thank the reviewers for their efforts and valuable feedback to improve our work.

\clearpage
%
%
\bibliographystyle{splncs04}
\bibliography{egbib}

\clearpage
\appendix
\section{Appendix}


\subsection{Qualitative Results}
\label{Sup:Qua}

\begin{figure}[h]
\centering
\includegraphics[width = 1\textwidth]{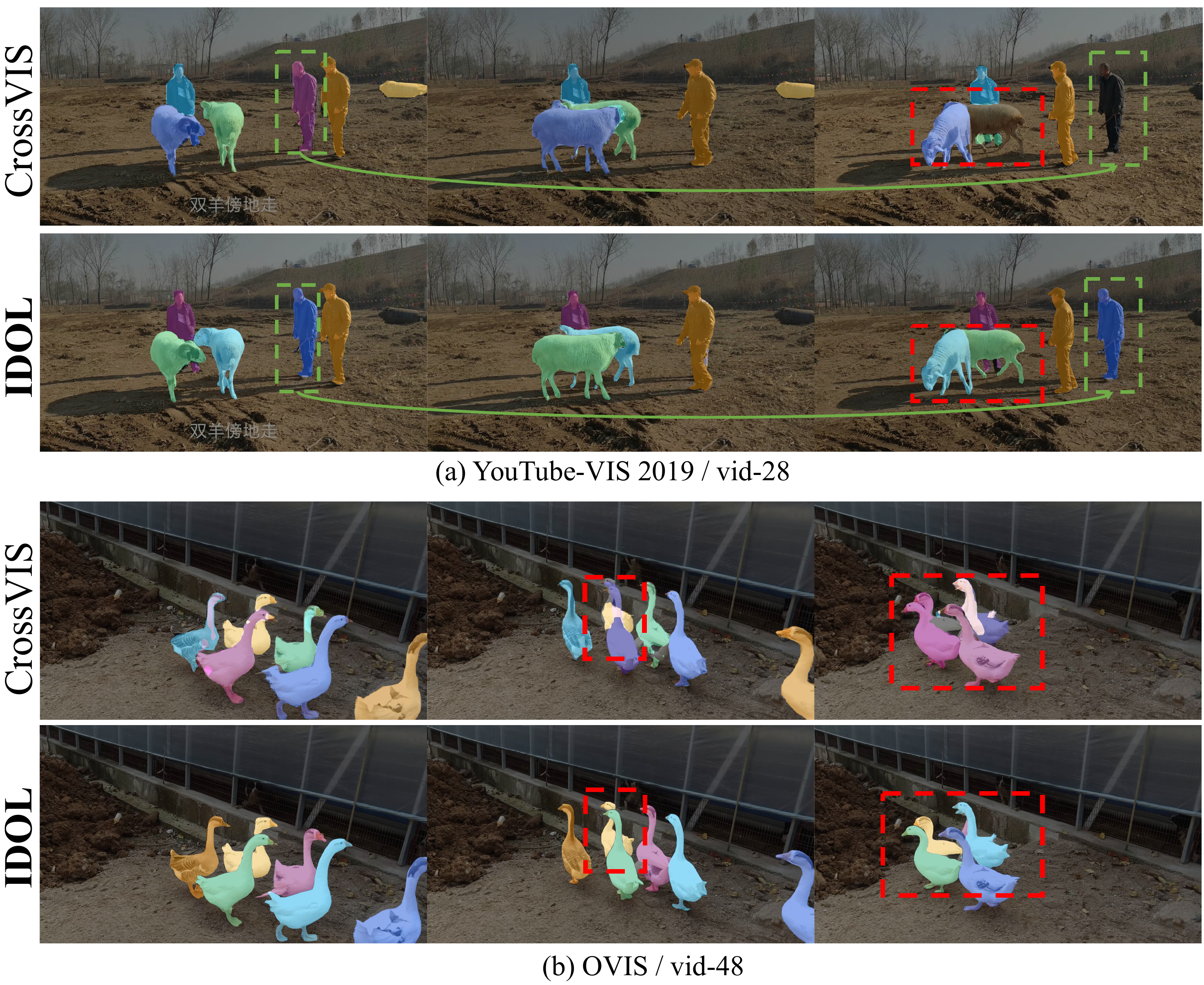}
\vspace{-2em}
\caption{Qualitative comparisons on videos with similar instances. Such kind of case is rare in YouTube-VIS, therefore we only select videos from OVIS. All methods use ResNet-50 backbone. Different color represents different instance id. Compare with the previous SOTA method, IDOL is able to segment and track instances with very similar appearances under complex motion and occlusions.}
\label{fig:compare_1}
\end{figure}

\begin{figure}[h]
\centering
\includegraphics[width = 1\textwidth]{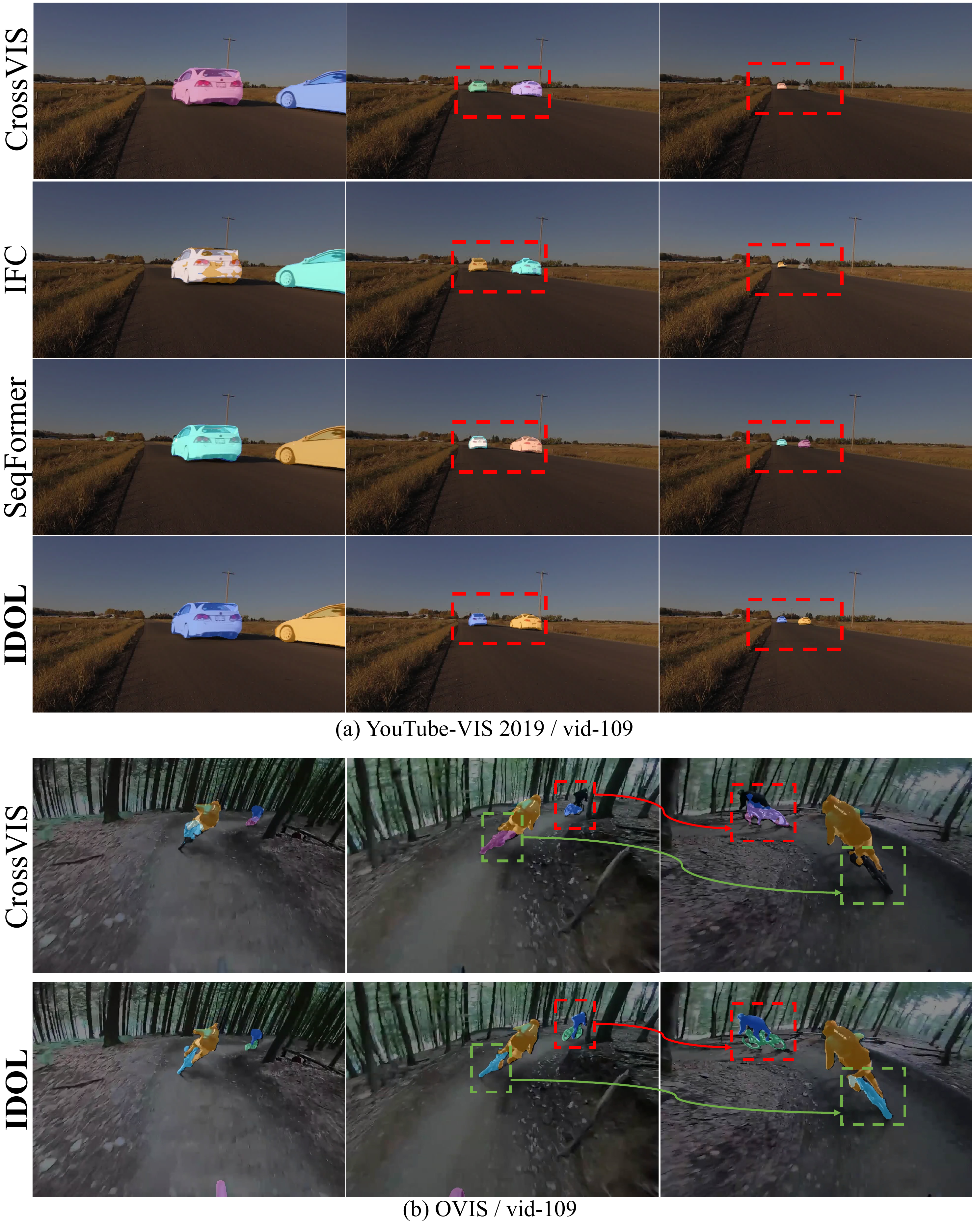}
\caption{Qualitative comparisons on videos with complex motions. We don't show the results of offline methods (IFC, SeqFormer) on OVIS since they do not provide official code/models on OVIS and the clip matching method provided by IFC fails in complex cases. All methods use ResNet-50 backbone. Different color represents different instance id. Compare with the previous SOTA methods, IDOL performs much better on videos with high-speed and large motions (a), and complex motions (b).}
\label{fig:compare_2}
\end{figure}

\begin{figure}[h]
\centering
\includegraphics[width = 1\textwidth]{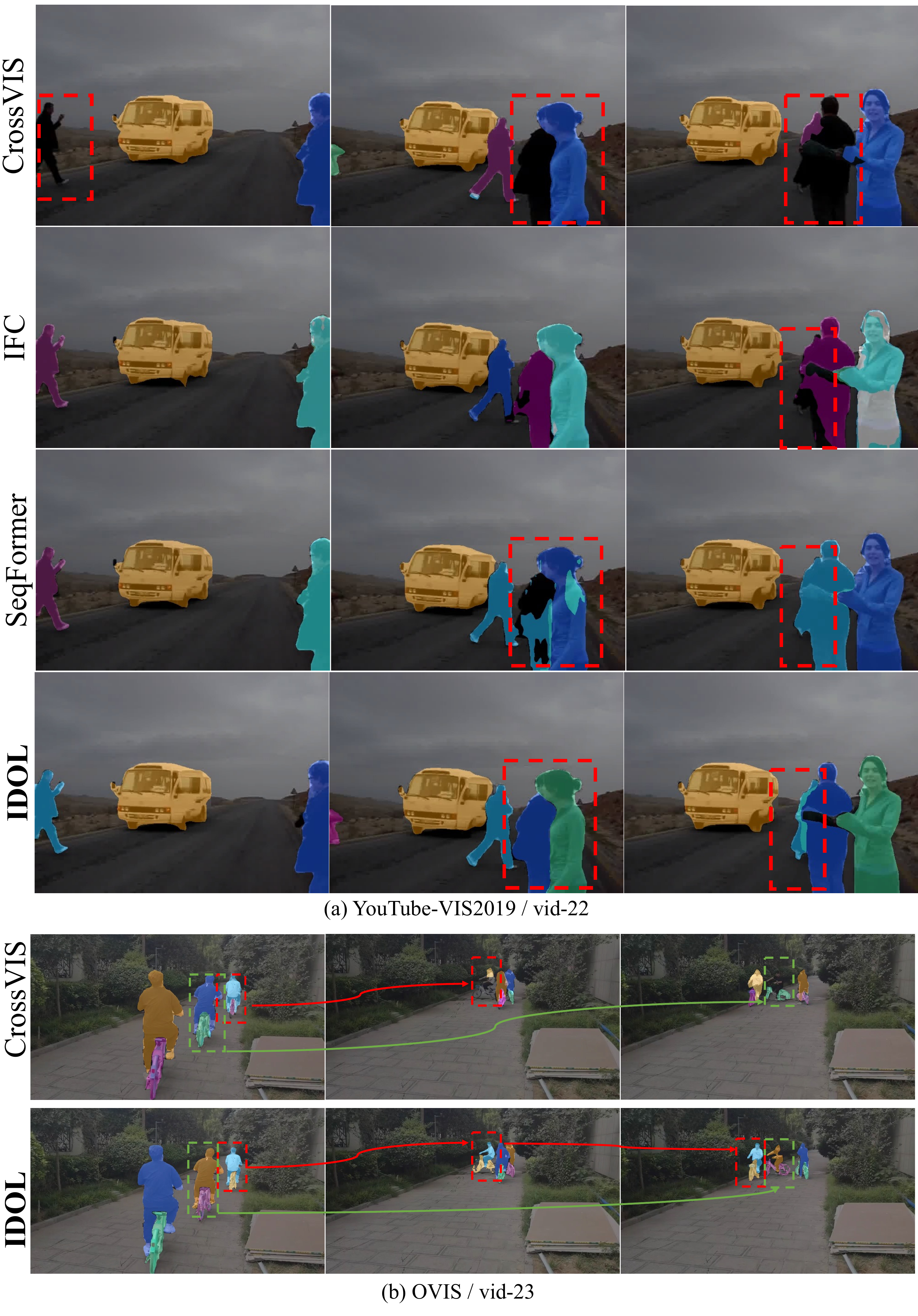}
\vspace{-2em}
\caption{Qualitative comparisons on videos with severe occlusions. All methods use ResNet-50 backbone. Different color represents different instance id. Compare with the previous SOTA methods, IDOL is more robust when handling crowded scenes with severe occlusions and frequent position exchanges.}
\label{fig:compare_3}
\end{figure}

In this section, we show several qualitative results on the validation sets of YouTube-VIS and OVIS to demonstrate the following advantages of IDOL:
\begin{itemize}

\item For instances that belong to the same category and have very similar appearances, our contrastive learning enables IDOL to segment and track these instances more accurately. (\eg Fig.~\ref{fig:compare_1})

\item Our method learns embedding with better temporal consistency, benefiting the tracking in videos with high-speed, large, and/or complex motions. (\eg Fig.~\ref{fig:compare_2})

\item With the help of more stable and discriminative embeddings, as well as our one-to-many temporally weighted softmax during inference, IDOL is more robust when handling crowded scenes with heavy occlusions and frequent position exchanges. (\eg Fig.~\ref{fig:compare_3})

\end{itemize}

\begin{figure}[h]
\centering
\includegraphics[width = 1\textwidth]{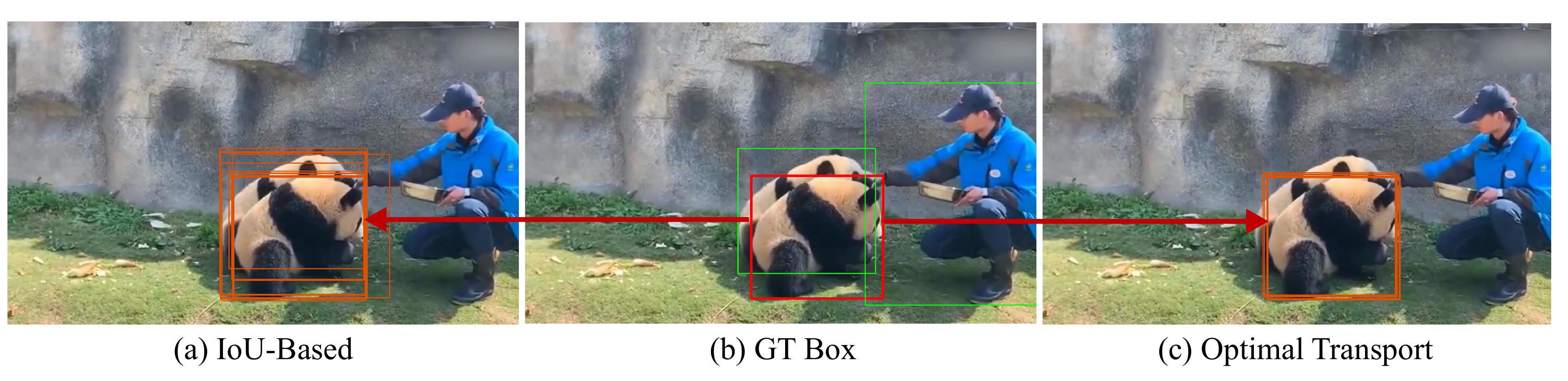}
\caption{Visualization of positive samples selected by IoU-based method (a) and our optimal transport method (c). The panda with red bounding box in (b) is the key instance. The positive samples selected by the IoU-based method are shown in (a), which causes false positives (\ie, the orange bounding box belonging to the panda behind the key instance). The positive samples selected by our method are shown in (c). It gives more accurate samples for positive embeddings and reduces false positives, further improving the quality of the embedding and the performance.
}
\label{fig:ot}
\vspace{-6mm}
\end{figure}

\subsection{Optimal Transport}
\label{Sup:opt}
Given a ground truth bounding box of an instance, the IoU-based method selects positive and negative samples by a hand-craft IoU threshold setting.
A predicted box is defined as positive to an instance if they have an IoU higher than 0.7, or negative if they have an IoU lower than 0.3, which introduces false positives in occlusions and crowded scenes.
As shown in Fig.~\ref{fig:ot} (a), 
in the case of occlusion between two pandas, IoU-based method would take the boxes belonging to the panda in the back as the positive samples of the front one, which causes false positives.
To address it, we formulate the problem of sample selection as an Optimal Transport problem in Optimization Theory, which reduces false positives and further improves the quality of the embedding.
For each ground truth, we sum the top 10 IoU values to get $m1$ and the top 100 IoU values to get $m2$. Then we take top $m1$ predictions with the lowest cost as positive and top $300-m2$ predictions with the highest cost as negatives. As shown in Fig.~\ref{fig:ot} (c), the optimal transport provides a better selection of positive embeddings during training, and thus improves the quality of the embedding.

\begin{figure}[h]
\centering
\includegraphics[width = 1\textwidth]{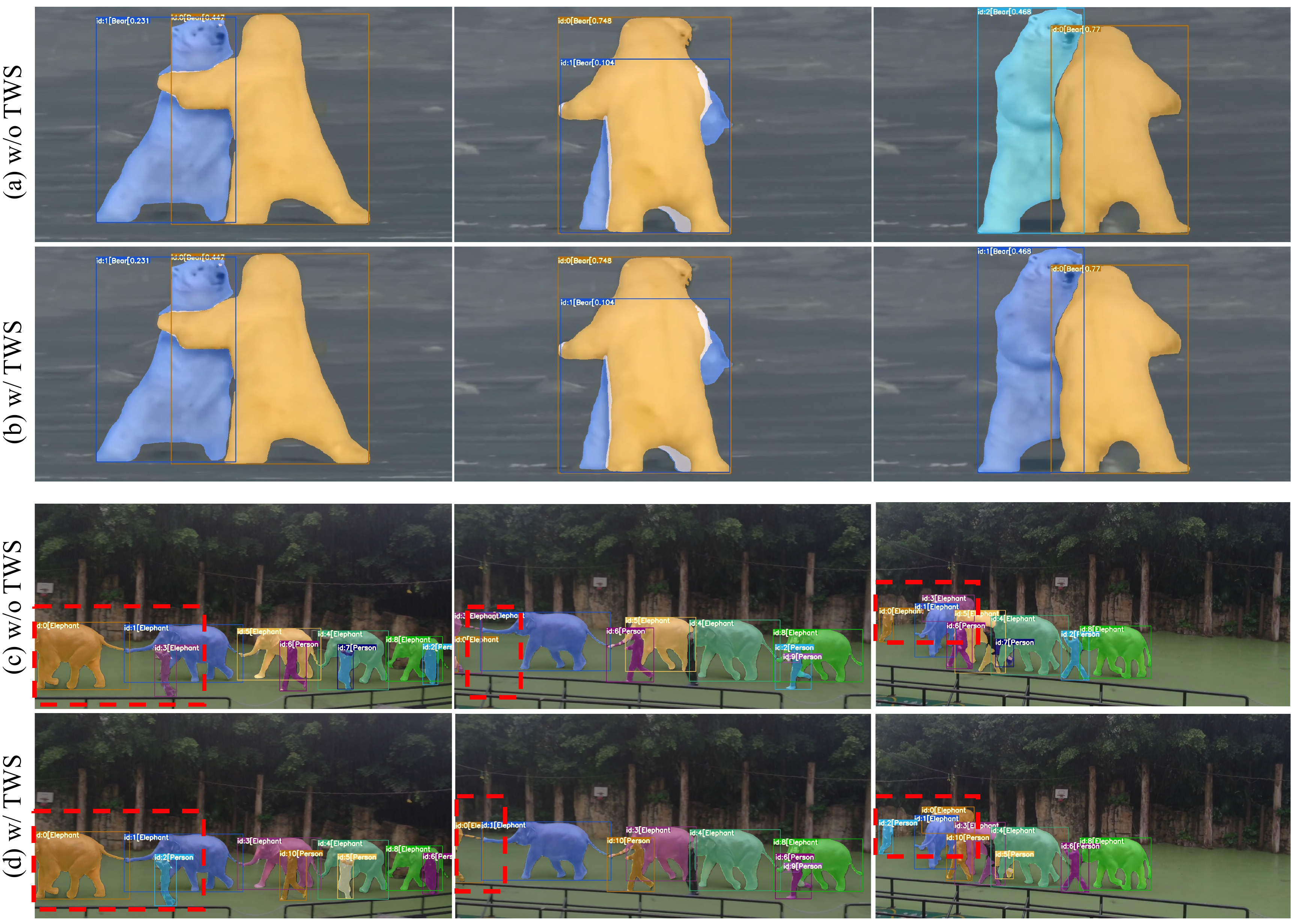}
\caption{Visualization of association quality with/without temporally weighted softmax (TWS).
Each row shows three adjacent frames from the same video. 
(a) and (c) show the association quality without temporally weighted softmax. (b) and (d) show the association quality with temporally weighted softmax.
The bear with `id:1' in (a) is occluded by another bear in some frames, and it is assigned a new id when it reappears. When the people with `id:3' and elephant with `id:0' in (c) disappear in the corner of the video and reappear after several frames, they are also assigned new ids. 
However, this problem is solved in (b) and (d) by our one-to-many temporally weighted softmax during inference.
}
\label{fig:temporal}
\vspace{-6mm}
\end{figure}

\subsection{Temporally Weighted Softmax}
\label{Sup:tem}
In Fig.~\ref{fig:temporal}, we show qualitative results of the temporally weighted softmax in our association strategy. 
As shown in Fig.~\ref{fig:temporal} (a) and (b), 
the bear with `id:1' in (a) is occluded by another bear in some frames, and without temporally weighted softmax, it is assigned a new id when it reappears. 
As shown in Fig.~\ref{fig:temporal} (c), the people with `id:3' and elephant with `id:0' disappear in the corner of the video, but they swap ids when they reappear after several frames, and this leads to classification errors.
However, in Fig.~\ref{fig:temporal} (d), temporally weighted softmax helps maintain temporal consistency of id for the sampe people and elephant.

\end{document}